\documentclass[11pt]{article}

\usepackage[preprint]{acl}

\usepackage{times}
\usepackage{latexsym}

\usepackage[T1]{fontenc}

\usepackage[utf8]{inputenc}

\usepackage{microtype}

\usepackage{inconsolata}

\usepackage{inconsolata}

\usepackage{graphicx}
\usepackage{booktabs}
\usepackage{multirow}
\usepackage{subcaption}
\usepackage[nameinlink,capitalize,noabbrev]{cleveref}

\usepackage{siunitx}  

\usepackage{tcolorbox}
\tcbuselibrary{listings}
\newtcolorbox{promptbox}[1][]{
  colback=gray!5,    
  colframe=gray!50,  
  fonttitle=\bfseries,
  coltitle=black,
  title=#1,
  boxrule=0.5pt,
  arc=2mm,           
  outer arc=2mm,
  left=4pt,right=4pt,top=4pt,bottom=4pt,
  after skip=12pt,
  listing only,
  listing options={
    basicstyle=\ttfamily\small,
    breaklines=true,
    columns=fullflexible
  }
}

\title{Perspectives on Cross-Lingual Consistency in LLMs for Medical Questions}

\author{\textbf{Minh Duc Bui$^{1}$}\quad~ \textbf{Mario Sanz-Guerrero$^{1}$}\quad~\textbf{Abteen Ebrahimi$^{2}$}\quad~\textbf{Sagi Shaier$^{1}$} \\
\textbf{Peter Herbert Kann$^{3}$}\quad~ \textbf{Manuel Mager$^{1, 4}$}\quad~ \textbf{Katharina von der Wense$^{1, 2}$} \\
\textsuperscript{1}Johannes Gutenberg University Mainz, Germany \quad \textsuperscript{2}University of Colorado Boulder, USA \\  \textsuperscript{3}University of Marburg, Germany \quad  \textsuperscript{4}Universidad Iberoamericana, Mexico \\
{\tt minhducbui@uni-mainz.de}}

\begin{document}
\maketitle

\begin{abstract}

Should multilingual LLMs answer medical questions consistently across input languages, or adapt responses to cultural cues? Existing multilingual medical benchmarks usually assume that medically correct answers should remain \textit{consistent} across languages and treat cross-lingual variation as model error. In contrast, cultural \textit{adaptation} research argues that appropriate medical answers may legitimately differ across contexts. We review the multilingual medical NLP literature through these two perspectives, we identify three gaps: limited stakeholder perspectives (e.g., of medical professionals), a lack of empirical evidence on which approach better serves users, and no benchmarks capable of distinguishing universally correct from culture-specific cases. To address the first gap, we survey 348 participants across three stakeholder groups (medical, NLP, and anthropology professionals) in three countries (Germany, Spain, and the United States). Anthropologists consistently favor adaptation, while medical and NLP respondents remain divided, with notable divergence between US and European medical professionals. LLMs prompted with profession and country personas fail to reproduce this variation, overestimating cross-lingual consistency preference among NLP and medical personas. We conclude that neither consistency nor adaptation can currently be considered clearly preferable, highlighting the need for empirical evidence on which approach better serves users across cultural contexts.

\end{abstract}

\section{Introduction}
\label{sec:intro}
Most multilingual medical benchmarks are built by translating questions and answers from a high-resource language, preserving the gold answer under translation and treating cross-lingual differences as model errors \cite{sviridova-etal-2024-casimedicos, matos-etal-2025-worldmedqa}. This reflects an assumption of \emph{consistency}: that correct answers are purely based on scientific knowledge and thus language-invariant---a view common in multilingual NLP, where language is treated as a neutral medium for conveying knowledge \cite{conneau-etal-2018-xnli, pmlr-v119-hu20b}.

\begin{figure}
    \centering
    \includegraphics[width=1.0\linewidth]{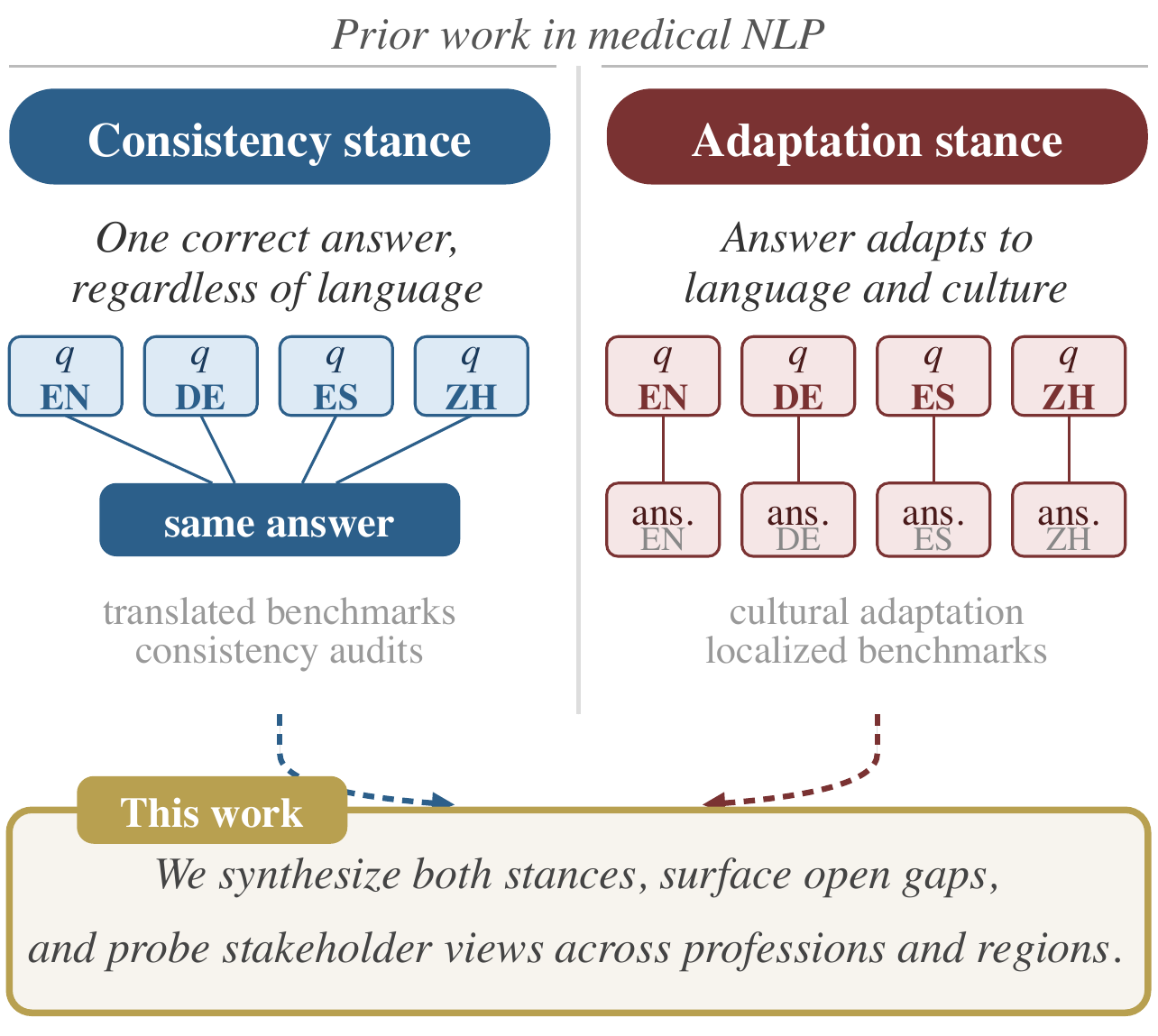}
    \caption{Prior work in multilingual medical NLP falls into two stances---\emph{consistency} and \emph{adaptation}---that rarely engage each other. We synthesize both, identify concrete gaps, and take a first empirical step toward evidence-based desiderata.}
    \label{fig:introduction}
\end{figure}

A second line of work rejects this assumption, arguing that, while language and culture are not interchangeable, they are entangled: they share common ground, values, and collective ways of making sense of the world \cite{hovy-yang-2021-importance, hershcovich-etal-2022-challenges, liu-etal-2025-culturally}. Under this view, correct answers may legitimately differ across languages, and \emph{adaptation}\footnote{Here \emph{adaptation} can mean \emph{content} (what is said) and \emph{communication} (how it is formulated).} to local norms is a feature rather than a failure. In medicine, the stakes are concrete: evaluations grounded in one cultural setting have been shown to penalize responses that are accurate and appropriate in another \cite{arora2025healthbenchevaluatinglargelanguage, nimo-etal-2025-africa}.

The two strands of thought (see Figure \ref{fig:introduction}) rarely engage  with each other's core assumption and, importantly, without finding a definite answer to the following important research question: \textit{should multilingual LLMs answer medical questions consistently across input languages, or should answers be adapted?}
In this paper, we start shedding light on this question via the following three contributions.

\paragraph{1) Literature Synthesis }

First, we synthesize the multilingual medical NLP literature into two competing stances: \emph{consistency}, which treats correct answers as language-invariant, and \emph{adaptation}, which treats them as partly contingent on cultural context. We survey prior works and their core arguments for each (Section~\ref{sec:review}). Comparing them surfaces three gaps that together define a research agenda (Section~\ref{sec:review-gaps}): (i) stakeholders are largely absent from the design loop; (ii) neither stance has been validated against downstream user benefit; and (iii) no benchmark distinguishes items that should be consistent (e.g., drug mechanisms) from those that may adapt (e.g., local guidelines).

\paragraph{2) Stakeholder Survey}

Second, we close the first gap through a survey of 348 participants across three stakeholder groups---medical, NLP, and anthropology professionals---and three countries: Germany, Spain, and the US (Section~\ref{sec:method}). Anthropologists lean toward adaptation; medical and NLP respondents are divided with no strong consensus for any stance. We additionally observe cross-national variation: US medical respondents slightly lean toward adaptation, while their German and Spanish counterparts do not. Taken together, these findings suggest that neither consistency nor adaptation can currently be considered the clearly superior approach, despite works in both camps expressing claims for their positions. This highlights the need for further investigation, particularly into user-centered approaches and suggests that the question may warrant the development of a dedicated research agenda of its own.

\paragraph{3) LLM Simulation of Disagreement}

Third, we investigate whether LLMs can accurately represent these stakeholder perspectives  (Section~\ref{sec:model}). If so, they could serve as a cost-effective proxy for stakeholder opinions. We find that they cannot: LLMs systematically overestimate support for the cross-lingual consistency stance among medical and NLP personas, and fail to reproduce the cross-national variation observed in our human survey.

\section{Background}

\subsection{Language vs. Culture}

Language and culture are deeply intertwined, yet one is rarely a reliable proxy for the other and conflating them is an error the NLP community has explicitly warned against \cite{hershcovich-etal-2022-challenges, adilazuarda-etal-2024-towards}. A question posed in Spanish may originate from a physician in Madrid, a community health worker in rural Oaxaca, or a patient in Buenos Aires---distinct cultural contexts that share a language for the most part but diverge substantially in health belief systems, clinical norms, and available resources. Treating language as a reliable index of culture obscures this variation.

Yet language remains the dominant proxy for culture in multilingual NLP \cite{arora-etal-2023-probing, naous-etal-2024-beer, bui-etal-2025-multi3hate}: it is always present, whereas richer cultural context---healthcare system, local guidelines, belief system---requires explicit specification that is frequently absent in practice. \citet{adilazuarda-etal-2024-towards} formalize this under the notion of \emph{proxies of culture}. When no such context is provided, language becomes the default carrier of cultural information by necessity rather than by design.

\subsection{Cultural Differences in Medical Practices} \label{sec:diff_practices}

Medical research data are increasingly accessible (e.g., via PubMed) and the international scientific community broadly agrees on classifying medical knowledge using evidence-based medicine criteria~\cite{sackett2000evidence}. Since clinical guidelines derive from this shared global evidence base, one might expect them to be identical worldwide. However, this is generally not the case.

Consider osteoporosis---a widely studied, epidemiologically important disease affecting millions worldwide. Comparing US guidelines~\cite{camacho2020osteoporosis} with those of German-speaking countries\footnote{\url{https://leitlinien.dv-osteologie.org}} reveals subtle but meaningful differences. Both aim to identify patients at high fracture risk requiring treatment, yet differ in their tools: the American guideline combines bone density with FRAX\textsuperscript{\textregistered}, a $\sim$10-variable fracture risk tool, whereas German-speaking countries employ a more granular tool incorporating bone density, gender, age, and $\sim$50 additional variables.

While not fundamental, these differences illustrate that guideline development is shaped by sociocultural factors, socioeconomic environments, and subjective preferences---not scientific evidence alone. This poses a key challenge for multilingual NLP systems: in medical contexts, there is rarely a single correct answer.

\section{Two Stances on Cross-Lingual Answer Consistency in Medical NLP}
\label{sec:review}

Existing work on multilingual and culturally adapted medical NLP offers two competing answers to the question of whether the response to a question should be consistent across languages or adapted to the linguistic and cultural context in which it is delivered. We now summarize the most important work for both stances, with the goal of identifying and highlighting the respective main arguments. We then discuss what the literature, taken together, leaves unresolved.

\subsection{The Consistency Stance}
\label{sec:camp1}
A first strand of work treats medical correctness as language-independent, i.e., content-wise identical questions in different languages should receive the same answer. This idea is found in two subareas: the \emph{construction} of multilingual benchmarks and the \emph{evaluation} of cross-lingual behavior.

\paragraph{Multilingual Medical Benchmarks} Most multilingual medical benchmarks are constructed by translating existing monolingual resources. \mbox{MedExpQA}~\cite{ALONSO2024102938} and \mbox{CasiMedicos}~\cite{GOENAGA2024102985} both extend a Spanish medical licensing exam into four languages, motivated by the lack of multilingual medical evaluation benchmarks. However, both retain the original Spanish physicians' answers as gold labels across all languages without consulting physicians from other countries---implicitly assuming that clinical correctness is invariant across languages. \mbox{JMedBench}~\cite{jiang-etal-2025-jmedbench} takes the same approach, treating English gold answers as Japanese ground truth, citing the insufficient scale of existing Japanese biomedical datasets. \mbox{XLingHealth}~\cite{Yiqiao} translates English gold-standard answers into other languages and frames consistency against them as a \emph{safety} criterion, stating that, in healthcare, ``incorrect or incomplete information can have life-threatening consequences,'' such that a model that deviates from the English answer in, e.g., Hindi is causing harm. The authors conclude that it is ``Better to Ask in English.'' Apollo's XMedBench \cite{wang2024apollolightweightmultilingualmedical} advances the \emph{language-neutral hypothesis}, suggesting an inclination to ``believe in the efficacy of multilingual training'' given that medical knowledge may be ``language-neutral to a significant extent,'' while acknowledging that multilingual training may undermine local specificities. Accordingly, its Hindi and Arabic splits are constructed by translating the English medical subset of MMLU \cite{hendryckstest2021}. This reasoning can be seen in the widespread practice of machine-translating MMLU~\cite{hendryckstest2021} for multilingual assessment~\citep{lai2023okapi,dang2024ayaexpanse,openai2024gpt4technicalreport,grattafiori2024llama3,bendale2024sutrascalablemultilingualmedical}. \citet{ferrazzi2026multilingualmedicalreasoningquestion} translates MedQA and MedMCQA into Italian and Spanish, offering no explicit justification.

Across these works, monolingual benchmarks implicitly function as the normative reference point: translated versions inherit the same gold answers, and deviations across languages are interpreted as model error rather than as potentially meaningful variation. The dominant motivation is the scarcity of multilingual medical resources~\citep{ALONSO2024102938,GOENAGA2024102985,jiang-etal-2025-jmedbench}, with translation treated as the most scalable remedy---rarely accompanied by deeper justification. Where such justification is offered, it tends toward a language-neutrality assumption: that medical knowledge is language-neutral~\citep{wang2024apollolightweightmultilingualmedical}.

\paragraph{Consistency Evaluation Across Languages } A parallel line of work tests whether deployed models give the same answer across languages, treating any divergence as a failure. \citet{10.1007/978-3-031-88714-7_30} document substantial inconsistency across English, German, Turkish, and Chinese on health-related questions, motivated by the observation that ``the quality of health information online varies by language, reflecting differences in national healthcare policies and practices'' and that ``non-English online health content often has lower quality.'' \citet{xu2026languageshapesmentalhealth} prompt GPT-4o and Qwen3 in Chinese and English on mental health tasks and find systematic shifts in both stigma judgments and depression severity classification, with Chinese prompts producing more stigmatizing outputs and underestimating severity relative to English. 

In each case, divergence from English or a language-independent reference response is treated as model failure.

\paragraph{Consistency Evaluation via Demographic Injection} Another line of work makes the consistency assumption explicit by inserting demographic information into otherwise identical prompts, all within a \textit{single language}, and treats any resulting change in model output as evidence of bias. \citet{shaier-etal-2023-emerging} curate biomedical questions intended to have ``demographically neutral answers'' and find that injecting patient demographic context frequently shifts model outputs, framing such changes as a fairness failure. DiversityMedQA \cite{rawat-etal-2024-diversitymedqa} perturbs MedQA items along gender and ethnicity, and \citet{rezaei2026counterfactualculturalcuesreduce} expand them by inserting cultural cues for different ethnicities, with a clinician verifying that the gold answer is consistent across variants.

Across all these works, however, consistency is assumed rather than proven. Since health systems, clinical communication norms, and treatment expectations differ across cultures, output changes are not always straightforwardly interpretable as bias, and the question of which language's answer is best (if any at all) remains open.

\subsection{The Adaptation Stance}
\label{sec:camp2}

A second stream argues for the claim that correct answers to medical questions are not universal across cultures.

\paragraph{Conceptual Foundations }
The foundation comes from outside the medical domain:
\citet{hershcovich-etal-2022-challenges} distinguish linguistic form, common ground,
aboutness, and values as separable dimensions of cross-cultural NLP, arguing
that benchmarks developed in one setting do not transfer cleanly to others.
\citet{liu-etal-2025-culturally} extend this with a fine-grained taxonomy of
culture and \citet{adilazuarda-etal-2024-towards} propose a typology of
demographic and semantic proxies of culture in LLMs. The medical works
below are domain-specific instantiations, organized along two axes: adaptation
in \emph{what} a model should say, and adaptation in \emph{how} it says it.

\paragraph{Native Resources}
A first claim concerns resource source: benchmarks should be built on native materials rather than translations. XMedBench \cite{wang2024apollolightweightmultilingualmedical}, while relying on translation for Hindi and Arabic, builds on native resources for its other four languages on the argument that ``local medical knowledge
can complement mainstream medical knowledge'' and ``improves communication efficiency and acceptance.'' MMedBench \cite{Qiu2024} also aggregates medical multiple-choice questions from examination banks across six languages, so each language's data is natively grounded. \mbox{Multi-OphthaLingua}~\cite{Restrepo} commissions board-certified native-speaker ophthalmologists to write questions in parallel across seven languages with explicit attention to question neutrality across regions.

\paragraph{Adaptation in Medical Content}
A second, stronger claim concerns medical knowledge itself: the correct answer can differ across settings. CMB \cite{wang-etal-2024-cmb} motivates a Chinese-localized benchmark by noting that a unified medical standard overlooks paradigms such as Traditional Chinese Medicine (TCM)---the authors warn explicitly that ``merely translating English-based medical evaluation may result in contextual incongruities to a local region.'' \citet{yizhen2024exploringcomprehensionchatgpttraditional} confirm that Western-trained models systematically lack the relevant TCM concepts and terminology.

Even within standard clinical practice, guidelines themselves vary by region, and \citet{zeng2025drllm} show that ChatGPT fails to reliably produce country-appropriate colorectal cancer screening recommendations, with the authors  highlighting its unreliability for geographically tailored clinical guidance. \citet{dey-etal-2025-beyond} evaluate a chatbot delivering sexual and reproductive health information to Indian women in local languages and find that culturally appropriate responses are systematically scored as incorrect against HealthBench \cite{arora2025healthbenchevaluatinglargelanguage}, whose rubrics encode Western clinical norms---for instance, penalizing India-specific dietary guidance because it did not match a US-market fish list, and rejecting locally grounded advice on breastfeeding because it omitted US federal law. The authors argue that such benchmarks ``overlook culture- and region-sensitivity'' and call for ``culturally adaptive evaluation frameworks that meet quality standards while recognizing needs of diverse populations.'' The Africa Health Check benchmark \cite{nimo-etal-2025-africa} similarly shows that medical LLMs default to allopathic, Western treatments even in zero-shot scenarios where local health systems would prescribe otherwise: the authors document a ``persistent default to allopathic (Western) treatments in zero-shot scenarios,'' and note that roughly 80\% of the African population relies on traditional herbal medicine for primary care---knowledge that current models largely ignore.

Lastly, \citet{calvo-bartolome-etal-2025-discrepancy} propose a user-in-the-loop fact-checking pipeline for detecting factual and cultural discrepancies in multilingual QA knowledge bases, taking a medical knowledge base as a case study, and explicitly acknowledging that two answers in different languages can each be grounded in credible evidence.

\paragraph{Adaptation of Communication }
A separate axis shifts the question from \emph{what} a model should say to \emph{how} it should say it: even where medical content is consistent, helpful and effective delivery might not be.\footnote{The CDC (a US national public health agency) explicitly recognizes that effective health communication is shaped by cultural factors; see \url{https://www.cdc.gov/health-literacy/php/develop-materials/culture.html}.} \citet{latam-cai-health-2025} run participatory workshops with citizens and health professionals across Latin America and argue that academic notions of culture lose meaning at the ground level: culturally appropriate conversational health AI needs to engage with how economics, infrastructure, geography, and local logistics are entangled with cultural experience, not just with language or surface social conventions. They propose a ``Pluriversal CAI in Health'' framework that centers relational context, including family as decision-making unit, community-level information flow, and material constraints on care, as central to the design problem rather than reducing cultural alignment to linguistic coverage alone. The medical NLP literature has so far engaged with this broader scope only sparsely.

\subsection{What the Literature Does Not Tell Us}
\label{sec:review-gaps}
After surveying the relevant work in the last two subsections, we make a couple of important observations, which we discuss below.

\paragraph{Stakeholder Absence from Design Loop } First, stakeholders are largely absent from the design loop. Most papers (for both lines of work) assume rather than measure the preferences of the populations a deployed multilingual medical LLM would serve. This omission matters because decisions about whether models should be consistent or culturally adaptive are not purely technical: they reflect value judgments about whose medical norms should prevail, and, without stakeholder input, those judgments risk encoding the researchers' assumptions rather than stakeholder expectations. Where stakeholders are consulted \cite{latam-cai-health-2025}, findings push toward a pluriversal framework foregrounding relationality and tolerance over data scaling, but samples remain geographically narrow, and focuses on health chatbots broadly rather than the specific question of cross-lingual consistency. No prior work has jointly surveyed both medical professionals and NLP researchers across multiple regions on whether multilingual medical LLMs should give the same answer in different languages. We address this gap in Section~\ref{sec:method}.

\paragraph{No User-Centered Outcome Evidence } Second, neither stance has empirically tested whether consistency or adaptation actually leads to better user outcomes. The consistency camp assumes consistency is more accurate; the adaptation camp assumes adaptation is more useful. But the entire debate is conducted at the level of benchmark construction and model evaluation, never at the level of downstream utility. Do patients in non-Western healthcare contexts make better decisions when a model adapts to local treatment norms or defers to international guidelines? \citet{dey-etal-2025-beyond} show that culturally appropriate responses are scored as incorrect by HealthBench, but do not assess how those responses serve the women who receive them. \citet{Yiqiao} conclude it is ``better to ask in English'' without investigating whether English answers are actually more useful to the people asking. The field therefore lacks basic evidence needed to adjudicate between the two stances: whether either one is more helpful for users.

\paragraph{Deciding between Consistency and Adaptation} Third, the two lines of work rarely engage with each other’s criticisms, and an important distinction is largely ignored: some medical knowledge transfers consistently across languages and regions (e.g., the mechanism of action of a drug), while other aspects of medical guidance legitimately vary across cultural, institutional, or national contexts (e.g., differing screening guidelines; see Section~\ref{sec:diff_practices}). The field of medical NLP still lacks 
research on which specific items consistency is expected or adaptation appropriate for, and it is still unknown to which degree NLP systems are able to make this distinction.

\begin{table}[t]
\centering
\small
\begin{tabular}{llr}
\toprule
\textbf{Culture} & \textbf{Profession} & \textbf{Number Participants} \\
\midrule
\multirow{3}{*}{Germany} & Anthropology & 17 \\
                         & Medical & 60 \\
                         & NLP & 53 \\
\midrule
\multirow{3}{*}{Spain} & Anthropology & 20 \\
                         & Medical & 57 \\
                         & NLP & 21 \\
\midrule
\multirow{3}{*}{USA} & Anthropology & 20 \\
                         & Medical & 53 \\
                         & NLP & 47 \\
\midrule
\multirow{3}{*}{\textit{Total}} & \textit{Anthropology} & \textit{57} \\
                                & \textit{Medical} & \textit{170} \\
                                & \textit{NLP} & \textit{121} \\
\bottomrule
\end{tabular}
\caption{
Number of participants per country/profession.}
\label{tab:counts}
\end{table}

\begin{figure*}[t]
    \centering
    \includegraphics[width=0.85\linewidth]{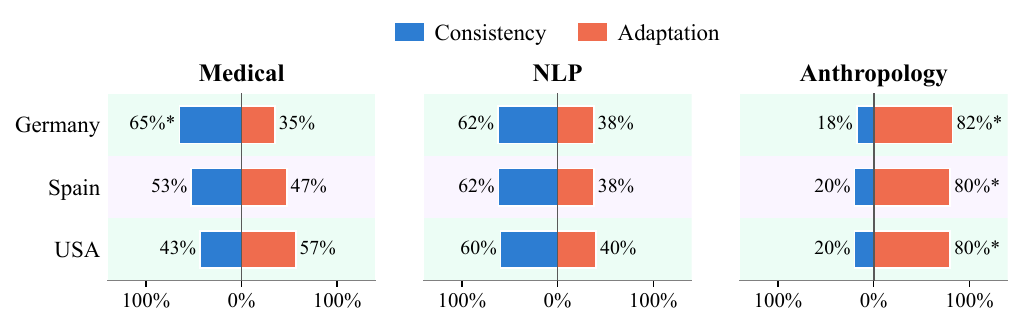}
    \caption{
    Responses by profession and country. Each bar shows the percentage of respondents preferring cross-lingual consistency (Answer~1) versus cultural adaptation (Answer~2). Asterisks (*) indicate significance according to a binomial test ($p < 0.05$) against a 50\% chance baseline.} 
    \label{fig:main_survey}
\end{figure*}

\section{Surveying Stakeholders}
\label{sec:method}
 
We address the first gap in Section~\ref{sec:review-gaps}---the lack of stakeholder involvement---by surveying professionals in medicine, NLP, and anthropology, testing whether the literature's consistency--adaptation tension extends to the stakeholders themselves.

\subsection{Target Population}
We sample along two axes: \textit{profession} and \textit{country}.

\paragraph{Profession} We target three professional groups. \textit{Medical professionals} would most directly use and be affected by deployed multilingual medical LLMs. \textit{NLP researchers} make design and evaluation decisions that shape whether deployed systems are consistent or adaptive. \textit{Anthropologists} are most attuned to how deeply culture shapes the way people understand, experience, and communicate.

\paragraph{Country} We recruit survey participants who are either citizens of or working in Germany, Spain, or the US---countries chosen to span distinct healthcare systems. This allows us to examine whether stakeholder views vary with institutional setting. The questionnaire is offered in English, Spanish, and German, so that respondents can complete it in the language of their choice.

\subsection{Recruitment}
We recruit through two channels: direct outreach via professional networks and mailing lists and the crowdsourcing platform Prolific for broader coverage. For Prolific participants, we used profession-based screening and attention checks for data quality; see Appendix \ref{ap:screening}. The final sample comprises 348 respondents; counts by profession and country are reported in Table~\ref{tab:counts}. Further, we report demographics related to the profession (e.g., year of graduation, or highest degree) in Appendix \ref{ap:demographics}.
 
\subsection{Survey}
The survey opens with a short framing paragraph noting that automated language systems sometimes return different answers to medical questions depending on the language in which the question is posed, even when the content is identical. Respondents are then asked which of two positions best reflects their own view: (1) that language should not influence the answer, which ought to be grounded in the current state of scientific knowledge (\emph{consistency}); or (2) that language is bound up with cultural identity, and that cultural identity carries distinct conceptions of illness, healing, and health---meaning answers should reflect the medical approaches prevailing within the cultural framework of the language used (\emph{adaptation}). The full questionnaire is provided in Appendix~\ref{ap:survey}.

The binary format is a deliberate methodological choice.\footnote{The questionnaire also included an \textit{other} option; these were manually mapped to one of the two stances where possible, or otherwise excluded ($<5\%$). We further analyze participants' optional free-text comments in Appendix~\ref{ap:freetext}.} By design, the two options are a close operationalization of the two stances we synthesize (Section~\ref{sec:review}), letting us map responses directly onto the debate as the literature frames it. Whether each stance applies at the level of individual items is a separate question, which we raise as an open gap (Section~\ref{sec:review-gaps}).

\subsection{Results and Discussion}
Figure~\ref{fig:main_survey} shows the survey participants' responses. In the following, we discuss the findings.

\paragraph{No Consensus among NLP and Medical Professionals }
Neither NLP nor medical professionals exhibit a strong preference in any country: the strongest agreement is for consistency, shared by 65\% of surveyed medical professionals in Germany. Still, a substantial minority favors adaptation in every group. A binomial test ($p<0.05$) against the 50\% chance baseline shows that 5 out of 6 profession--country combinations do not differ significantly from chance, with the German medical group being the only exception---highlighting the need to explicitly tackle the question.


\paragraph{Anthropologists Favor Adaptation }
Anthropologists are the only group to exhibit a clear and consistent preference, favoring adaptation in all three countries with support never falling below 75\%. All three country-level results differ significantly from the 50\% chance baseline (binomial test, \(p < 0.05\)). The stability of this preference across substantially different healthcare contexts is interpretively straightforward: anthropologists are trained to treat cultural variation as the default condition of human life and to be suspicious of universalizing claims.

\begin{table}[t]
\centering
\small
\begin{tabular}{lcc}
\toprule
 & \multicolumn{2}{c}{Pro Consistency (\%)} \\
\cmidrule(lr){2-3}
Profession & Humans & LLMs \\
\midrule
Anthropology & $19.2{\scriptstyle\,\pm\,1.4}$ & $16.2{\scriptstyle\,\pm\,3.7}$ \\
NLP & $61.2{\scriptstyle\,\pm\,1.5}$ & $80.0{\scriptstyle\,\pm\,3.9}$ \\
Medical & $53.7{\scriptstyle\,\pm\,10.8}$ & $83.6{\scriptstyle\,\pm\,4.1}$ \\
\bottomrule
\end{tabular}
\caption{Proportion selecting consistency per profession for humans (averaged across countries) and LLMs (averaged across countries, languages and models).}
\label{tab:consistency_profession}
\end{table}

\paragraph{Support of the Consistency Stance among Medical Professionals Is Weakest in the US }
German and Spanish medical respondents show a tendency toward consistency, whereas their US counterparts do not: 57\% of US medical respondents select adaptation, as compared to 35\% in Germany and 47\% in Spain. A two-sided proportion z-test indicates a significant difference between the US and Germany ($p < 0.05$). One plausible explanation is that awareness of cultural differences in medical practice is more prevalent in the US---liability culture, insurance structures and clinical guidelines \cite{office2001national}.

 \begin{figure}
    \centering
    \includegraphics[width=1.0\linewidth]{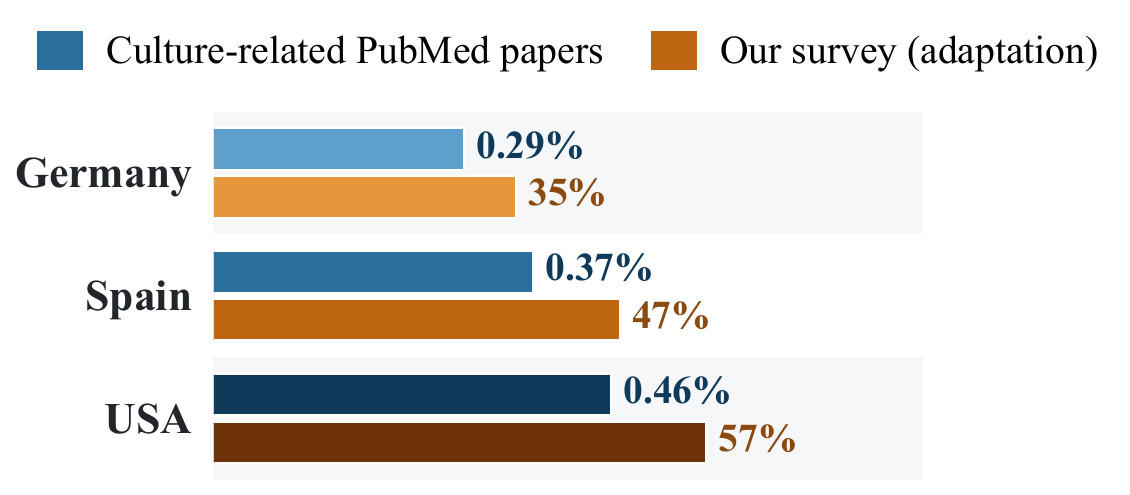}
    \caption{
    The proportion of PubMed articles from 2015 to 2025 affiliated with each country that carry at least one MeSH term from the \textit{Culture} subtree. This is compared against the percentage of responses by medical professionals in favor of adaptation (see Figure \ref{fig:main_survey}).}
    \label{fig:pubmed}
\end{figure}

To substantiate this empirically, we analyze PubMed publications (2015--2025) from all three countries; see Appendix~\ref{ap:pubmed} for detailed setup and results. We find that the country ranking mirrors our survey ranking (see Figure \ref{fig:pubmed}): US publications show the highest share of culture-related articles, followed by Spain and Germany, lending empirical support to the view that awareness of cultural differences in the US shapes not only clinician preferences but the broader biomedical research agenda.

\subsection{Implications of the Survey Findings}
Taken together, these findings suggest that neither consistency nor adaptation commands clear consensus, despite advocates on both sides expressing confident claims. This highlights the need for further investigation, particularly into user-centered approaches as suggested by our literature synthesis (see Section \ref{sec:review-gaps}), and suggests that the question may warrant the development of a dedicated research agenda of its own.

Notably, this absence of consensus emerges already \emph{within} three Western, Global-North systems, alongside a significant US--Germany difference. In settings where traditional or alternative medicine plays a larger role in care, such as contexts shaped by traditional Chinese medicine \citep{wang-etal-2024-cmb}, support for adaptation is plausibly stronger still, so a geographically broader sample would likely reinforce rather than overturn our central finding.

\begin{figure*}[t]
    \centering
    \begin{subfigure}{\linewidth}
        \centering
        \includegraphics[width=0.9\linewidth]{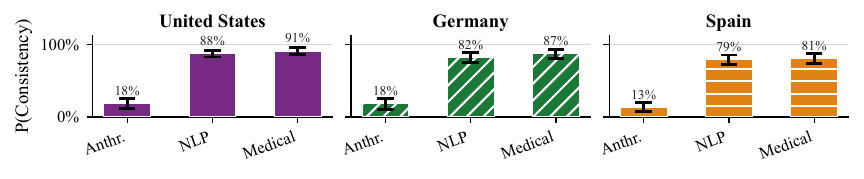}
        \caption{\textbf{By Country (English Prompts).} Proportion selecting consistency, averaged across models.}
        \label{fig:country_bias}
    \end{subfigure}
    
    
    \begin{subfigure}{\linewidth}
        \centering
        \includegraphics[width=0.9\linewidth]{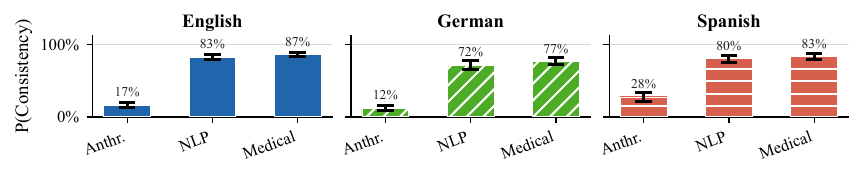}
        \caption{\textbf{By Prompt Language.} Proportion selecting consistency, averaged across models.}
        \label{fig:language_bias}
    \end{subfigure}
    \caption{Percentage of answers in favor of consistency, averaged across models, conditioned on country persona (top) and prompt language (bottom).}
    \label{fig:country_language_bias}
\end{figure*}

\section{LLM Simulation of Survey Opinions} \label{sec:model}
Based on our survey data, we now ask the question of whether LLMs accurately represent the opinions of different professional and country groups regarding the survey question. If that were the case, they would be useful tools, e.g., for estimating stakeholder opinions before deciding on conducting additional or in-depth surveys, saving both time and money. For this, we examine how LLMs respond to the same survey question posed to human participants when provided with relevant personas.

\subsection{Experimental Setup}

\paragraph{Models } We evaluate 11 open-weight instruction-tuned models spanning a range of sizes and model families: Gemma~4 (E4B, 26B-A4B, 31B), Llama~3.1 \citep[8B;][]{grattafiori2024llama3}, Llama~3.3 (70B), Qwen3.5 \citep[9B, 27B, 35B-MoE;][]{qwen35blog}, Phi-4 \citep[14B;][]{abdin2024phi4}, Aya Expanse \citep[32B;][]{dang2024ayaexpanse}, and GPT-OSS \citep[20B;][]{openai2025gptoss}.

\paragraph{Setup} Each model receives the survey in one of three languages (English, German, Spanish) under one of three profession personas (medicine, NLP, anthropology) paired with three countries (US, Germany, Spain), conveyed via the system prompt~\cite{zheng-etal-2024-helpful}. To control for position bias, answer options are randomly shuffled across runs. Each condition is repeated $n{=}20$ times with sampling-based decoding (temperature $= 0.7$, top-$p = 0.9$). We report the proportion selecting the consistency option, averaged across models. Prompt details and hardware are in Appendix~\ref{ap:setup_details}.

\subsection{Results}

Detailed results for all combinations, including LLMs without personas, are in Appendix~\ref{ap:model_results_detailed}; below we aggregate results for a concise overview.

\paragraph{LLMs Overestimate Approval of the Consistency Stance among NLP and Medical Professionals}
Table~\ref{tab:consistency_profession} compares the proportion of human and LLM respondents selecting the consistency option. While, at the aggregate level, humans are roughly evenly split, LLMs exhibit a stronger lean toward consistency under NLP and medical personas ($80.0\%$ and $83.6\%$, respectively), compared to $61.2\%$ and $53.7\%$, respectively, among humans. Thus, LLMs do not accurately reflect the opinions of human stakeholders in 2 out of 3 cases.

\paragraph{LLMs Lack the Country-Level Variation Observed in Humans }
An interesting finding among human respondents is that the opinions regarding consistency among medical professionals differ across countries: US medical respondents favor adaptation ($57\%$), while their German and Spanish counterparts lean toward consistency ($35\%$ and $47\%$ for adaptation, respectively). LLMs show no analogous sensitivity to country persona---agreement with the consistency stance remains high across countries (Figure~\ref{fig:country_bias}), suggesting that system prompts fail to induce the variation that exists in human expert judgment.

A similar pattern is found across languages (Figure~\ref{fig:language_bias}): English prompts yield the highest consistency rates for NLP and medical personas ($0.80$ and $0.84$), while German prompts produce a notable drop ($0.67$ and $0.69$), with Spanish falling in between ($0.74$ and $0.79$). This is noteworthy given that, in the human survey, participants were surveyed in their native language (or that of their workplace)---yet US medical respondents (surveyed in English) showed the \emph{lowest} consistency preference among medical professionals. LLMs thus exhibit the opposite pattern: English prompts push models \emph{toward} consistency, whereas the language most associated with high human consistency (German) nudges models toward adaptation.

\subsection{Practical Implications}

First, LLMs given a persona cannot stand in for real stakeholder surveys when making design choices: they overrate support for the consistency stance among NLP and medical personas (Table~\ref{tab:consistency_profession}) and do not show the cross-country variation we find in humans (Figure~\ref{fig:country_bias}). Furthermore, models lean toward consistency by default, most strongly under English prompts (Figure~\ref{fig:language_bias}), so deployed multilingual medical systems likely carry this same bias.

Secondly, for building such systems, the next question is how to steer a model toward the behavior we want---but the field first has to decide which behavior that is (Section~\ref{sec:review-gaps}). Once a target is set, (i) consistency could be pursued through cross-lingual alignment on parallel medical QA with shared answers, and (ii) adaptation through natively sourced per-region QA and retrieval over country-specific clinical guidelines. Our results caution, however, that system-prompt steering alone fails to reproduce even coarse stakeholder variation, so prompting is unlikely to be a sufficient adaptation mechanism.

\section{Conclusion}

This work synthesizes two competing stances on cross-lingual consistency in medical NLP and identifies three gaps: stakeholders (e.g., medical professionals) are absent from the design loop, no study has tested whether either stance improves user outcomes, and no benchmark separates items that should be consistent from those that should adapt. Our stakeholder survey addresses the first gap: anthropologists show strong consensus for adaptation; medical professionals and NLP researchers are divided. We also observe a notable cross-country shift among US-based medical respondents toward adaptation. Our LLM evaluation shows that current models fail to reproduce stakeholder distributions or replicate this cross-country variation. Taken together, neither stance commands clear consensus---despite claims on both sides---highlighting the need for user-centered empirical investigation.


\section*{Limitations}

The survey (Section \ref{sec:method}) was designed to answer one question: whether systematic stakeholder disagreement or agreement exists. It is appropriately scoped for that question rather than for others. Three limitations are worth making explicit. First, the binary instrument cannot tell us \emph{which dimensions} of the consistency--adaptation distinction respondents weighted in their answer: medical content, communication style, or both. We chose the forced choice deliberately, for the reasons given in Section \ref{sec:method}, but a follow-up study should decompose the question along these axes. Second, sample sizes within individual profession-by-country groups are adequate to surface the patterns we report but not to support fine-grained statistical comparison across demographics. Third, our sample is drawn entirely from the Global North, which limits the generalisability of our findings and leaves open whether the patterns we observe hold in low- and middle-income country contexts, where multilingual medical AI may ultimately have the greatest impact.

\section*{Ethical Statement}

The crowdworkers were recruited through Prolific and compensated at a rate of \$16.60 per hour, which exceeds the minimum wage in the authors' country, ensuring fair payment. The project received IRB ethical approval before annotations began. Annotators were thoroughly informed about the project's content and purpose and gave explicit consent before starting. No personally identifiable information was collected.

We acknowledge the potential risks of our work, including that our findings could be misused to justify medically harmful LLM outputs under the guise of adaptation; however, we believe that openly surfacing this tension and calling for further research is a necessary step toward responsible development of multilingual medical AI.

We use AI writing assistants solely for sentence-level editing.

\section*{Acknowledgments}

This work was supported by the Carl Zeiss Foundation through the TOPML and MAINCE projects (grant numbers P2021-02-014 and P2022-08-009). We also thank all survey participants and everyone who helped distribute it through their networks and mailing lists.

\bibliography{anthology-1, anthology-2, refs}

\appendix

\section{Survey Details}
\label{ap:survey}

\subsection{Filtering Crowdworkers} \label{ap:screening}

To ensure that crowdworkers belong to the target professions, we apply a three-stage filtering pipeline:
(1)~Prolific pre-screening, (2)~survey-level domain filtering, and (3)~a profession-specific
verification question (applied only to NLP researchers and anthropologists).

For \textbf{medical professionals}, Prolific provides a dedicated filter that restricts participants
to occupations within medicine. For \textbf{NLP researchers}, Prolific does not offer a
domain-specific filter; we therefore pre-screen for participants employed in the
\textit{Information Technology} sector. Similarly, for \textbf{anthropologists}, we pre-screen
within the \textit{Social Sciences} sector.

In the second stage, all participants are asked within the survey itself to indicate their
domain of expertise---NLP, Anthropology, or Medicine---to confirm alignment with the target group.

In the third stage, we apply profession-specific verification questions to further validate
expertise. NLP researchers are asked to complete the fill-in-the-blank prompt
\textit{``Attention is all you \texttt{[MASK]}''},\footnote{The correct completion is
\textit{need}, referencing the seminal transformer paper by \citet{NIPS2017_3f5ee243}.}
while anthropologists are asked to mention their specific subfield in free text.
All responses are subsequently reviewed manually to confirm eligibility.

\subsection{Survey Form}

Figure~\ref{fig:survey_question} shows the main survey question presented to all participants.

\begin{figure}[t]
    \centering
    \begin{promptbox}[Survey Question]
        \ttfamily\small\raggedright
        When an automated system (artificial intelligence, language model) is asked about a medical issue---such as a diagnosis, a recommended therapy, or the interpretation of findings---some answers differ depending on the language in which the question is asked (e.g., German, French, English, Spanish, Indigenous languages in South America, Africa, or Asia). This occurs even when the content of the question is completely identical.\\[0.5em]
        Which of the following opinions would you be more inclined to support:\\[0.5em]
        1) The language in which the question is asked should not influence the answer, as it should be based exclusively on the current state of scientific knowledge.\\
        2) Language is part of cultural identity. Cultural identity also implies different concepts of illness, healing, and health. Accordingly, the answer should refer to the medical practices applicable within the cultural framework to which the language of the question belongs.\\
      Other: <Free Text Form>
    \end{promptbox}
   \caption{The main survey question shown to all participants. Participants were asked to
    assess a given statement and provide a response.}
    \label{fig:survey_question}
\end{figure}

The translated versions of the survey were verified by native speakers from each country.

\subsection{Demographic Details} \label{ap:demographics}

To make our survey as light as possible, we decided to only collect demographics related to profession. We report these in the Table \ref{tab:demographics}.

\begin{table*}[t]
\centering
\small
\setlength{\tabcolsep}{4pt}
\begin{tabular}{llrrrrrrrrrrrr}
\toprule
\textbf{Culture} & \textbf{Job} & & \multicolumn{4}{c}{\textbf{Graduation Year}} & \multicolumn{7}{c}{\textbf{Degree}} \\
\cmidrule(lr){4-7} \cmidrule(lr){7-13}
 & & \textbf{N} & \textbf{Mean} & \textbf{SD} & \textbf{Min} & \textbf{Max} & \textbf{HS} & \textbf{Voc.} & \textbf{Bach.} & \textbf{Mast.} & \textbf{PhD} & \textbf{Hab.} & \textbf{Other} \\
\midrule
\textit{Germany} & Anthropology & 17 & 2017.3 & 8.4 & 1991 & 2025 & -- & -- & 4 & 4 & 6 & 3 & -- \\
 & Medical & 60 & 2019.1 & 8.4 & 1982 & 2030 & 9 & 8 & 18 & 15 & 10 & -- & -- \\
 & NLP & 53 & 2020.7 & 5.7 & 1997 & 2027 & 3 & 3 & 9 & 27 & 11 & -- & -- \\
\midrule
\textit{Spain} & Anthropology & 20 & 2017.0 & 13.0 & 1978 & 2026 & 3 & 4 & 6 & 1 & 5 & 1 & -- \\
 & Medical & 57 & 2014.2 & 12.3 & 1984 & 2026 & 2 & -- & 10 & 20 & 21 & 3 & 1 \\
 & NLP & 21 & 2020.0 & 5.9 & 1998 & 2026 & -- & 5 & 7 & 7 & 2 & -- & -- \\
\midrule
\textit{USA} & Anthropology & 20 & 2017.7 & 11.9 & 1973 & 2026 & -- & 1 & 5 & 12 & 1 & 1 & -- \\
 & Medical & 53 & 2014.9 & 10.3 & 1977 & 2029 & 7 & 6 & 22 & 13 & 3 & 1 & 1 \\
 & NLP & 47 & 2012.3 & 12.0 & 1970 & 2029 & 7 & 6 & 17 & 11 & 3 & -- & 2 \\
\bottomrule
\end{tabular}
\caption{Participant demographics by culture and job role. Graduation Year: N=count, SD=standard deviation, Min/Max=range. Degree counts: HS=High School, Voc.=Vocational, Assoc.=Associate, Bach.=Bachelor, Mast.=Master, Hab.=Habilitation/Postdoctoral.}
\label{tab:demographics}
\end{table*}

\subsection{Free-Text Responses} \label{ap:freetext}

The survey ended with an optional field in which participants could elaborate on their view. Of the 348 respondents, 60 left a comment, of which 44 were relevant; the remaining 16 were acknowledgements, contact requests, or non-answers (e.g., ``N/A''). We read the relevant comments inductively and grouped them into six recurring themes, assigning each comment to its primary theme. Because the responses are few and some touch on more than one theme, we treat the counts in Table~\ref{tab:freetext} as indicative rather than exact.

The most frequent theme reproduces the content--communication distinction of Section~\ref{sec:camp2}: respondents hold that the medical \emph{content} should stay consistent across languages while its \emph{communication} may adapt to the reader (e.g., ``the medically relevant information should be phrased language-independently\ldots but \emph{how} it is communicated may well depend on cultural aspects'').

\begin{table}[t]
\centering
\small
\begin{tabular}{lr}
\toprule
\textbf{Theme} & \textbf{\#} \\
\midrule
Keep content, adapt communication & 13 \\
Illness is culturally situated & 8 \\
Language $\neq$ culture & 8 \\
Adaptation risks harmful bias & 7 \\
Translation and terminology risk & 5 \\
Critique of the binary instrument & 3 \\
\bottomrule
\end{tabular}
\caption{Themes in the free-text comments. Each comment is assigned its primary theme.}
\label{tab:freetext}
\end{table}

\section{PubMed Crawling}
\label{ap:pubmed}

We crawled PubMed articles published between 2015 and 2025, following the procedure described in the \texttt{ncbi/pubmed} dataset documentation from Hugging Face. For each article, we extracted author affiliations and retained only publications in which all authors were affiliated with institutions from the same country.

To identify culturally relevant medical literature, we selected articles associated with MeSH terms related to culture from the categories \textit{Anthropology, Cultural} and \textit{Sociological Factors} (e.g., \textit{Cross-Cultural Comparison}, \textit{Acculturation}). These filters were used to construct the culturally grounded subset analyzed in our experiments.

\begin{figure}[t]
    \centering
    \includegraphics[width=1.0\linewidth]{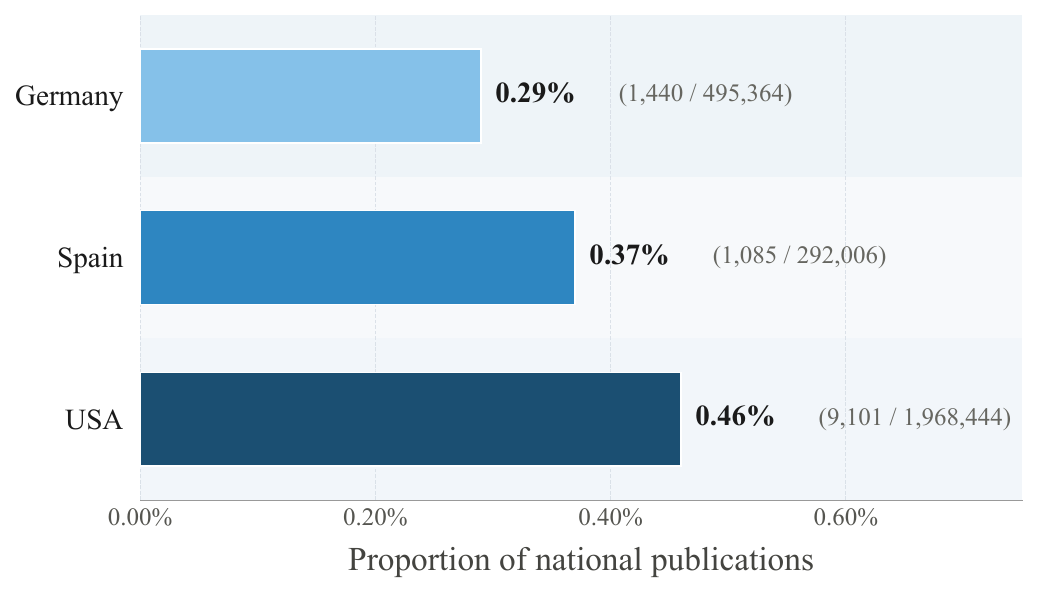}
    \caption{Detailed Results for PubMed Publications.}
    \label{fig:detail_pubmed}
\end{figure}

We report the comparison to our survey in Figure \ref{fig:pubmed} and detailed results in Figure \ref{fig:detail_pubmed}.

\section{Detailed Model Experiments}

\begin{figure}[t]
    \centering
    \begin{promptbox}[Prompt Template]
        \ttfamily\small\raggedright
        \textbf{\{system\_token\}}\\
        ``You are a person \{profession\} currently working in \{country\}. Be short. Answer the following survey question honestly and from your professional perspective.''\\[0.5em]

        \textbf{\{user\_token\}}\\
        ``\{survey text\}\\
        1) \{option\_1\}\\
        2) \{option\_2\}\\
        Please respond with either `(1)' or `(2)', followed by a brief explanation of your choice.''\\[0.5em]
    \end{promptbox}
    \caption{Prompt template used for the survey experiments. We provided the same prompt to the model in different languages (English, German, and Spanish). The order of the two response options was randomized for each prompt instance to mitigate positional bias.}
    \label{fig:survey_llm_prompt}
\end{figure}

\subsection{Setup Details} \label{ap:setup_details}

The full prompt is provided in Figure \ref{fig:survey_llm_prompt}. All experiments were conducted on 2 H100 GPUs; a single LLM inference pass takes approximately 2 minutes for the largest models.

\subsection{Detailed Results} \label{ap:model_results_detailed}

We report the model results without persona in Table \ref{tab:default_consistency_languages}. We report the detailed model results with persona in Table \ref{tab:consistency_all}.

\begin{figure}[t]
    \centering
    \includegraphics[width=0.9\linewidth]{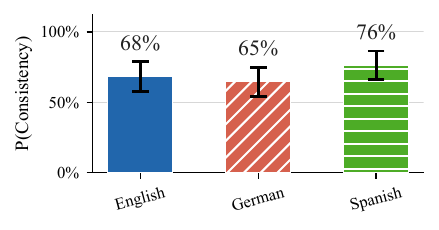}
    \caption{\textbf{No Persona Result.} Proportion of selecting consistency, averaged across models.}
    \label{fig:profession_consistency}
\end{figure}

\paragraph{LLMs without Personas} We find weak consensus without personas, see Figure~\ref{fig:profession_consistency}. Averaged across LLMs, responses exhibit substantial variance and only a weak consensus, with a slight overall preference for the consistency option.

\begin{table}[t]
\centering
\small
\setlength{\tabcolsep}{4pt}
\begin{tabular}{lrrr}
\toprule
\textbf{Model} & \textbf{English} & \textbf{German} & \textbf{Spanish} \\
\midrule
Llama-3.1-8B-Instruct & 6/14 & 7/13 & 3/17 \\
Llama-3.3-70B-Instruct & 20/0 & 13/7 & 20/0 \\
Qwen3.5-27B & 20/0 & 20/0 & 20/0 \\
Qwen3.5-35B-A3B & 17/3 & 15/5 & 18/2 \\
Qwen3.5-9B & 14/6 & 16/4 & 18/2 \\
aya-expanse-32b & 0/20 & 0/20 & 20/0 \\
gemma-4-26B-A4B-it & 20/0 & 20/0 & 20/0 \\
gemma-4-31B-it & 20/0 & 20/0 & 20/0 \\
gemma-4-E4B-it & 18/2 & 4/16 & 3/17 \\
gpt-oss-20b & 8/12 & 11/9 & 11/9 \\
phi-4 & 7/13 & 16/4 & 15/5 \\
\bottomrule
\end{tabular}
\caption{Consistency results (Consistent/Adaptation) with no persona across languages.}
\label{tab:default_consistency_languages}
\end{table}

\begin{table*}[t]
\centering
\tiny
\begin{tabular}{lrrrrrrrrr}
\toprule
\textbf{Model} & \multicolumn{3}{c}{\textbf{Germany}} & \multicolumn{3}{c}{\textbf{Spain}} & \multicolumn{3}{c}{\textbf{United States}} \\
\cmidrule(lr){2-4}
\cmidrule(lr){5-7}
\cmidrule(lr){8-10}
 & \textbf{Anthr.} & \textbf{NLP} & \textbf{Medical} & \textbf{Anthr.} & \textbf{NLP} & \textbf{Medical} & \textbf{Anthr.} & \textbf{NLP} & \textbf{Medical} \\
\midrule
\multicolumn{10}{l}{\textbf{Language: English}} \\
\midrule
Llama-3.1-8B-Instruct & 6/14 & 14/6 & 7/13 & 3/17 & 10/10 & 12/8 & 8/12 & 14/6 & 19/1 \\
Llama-3.3-70B-Instruct & 0/20 & 20/0 & 20/0 & 0/20 & 20/0 & 20/0 & 0/20 & 20/0 & 20/0 \\
Qwen3.5-27B & 8/12 & 20/0 & 20/0 & 1/19 & 17/3 & 20/0 & 11/9 & 20/0 & 20/0 \\
Qwen3.5-35B-A3B & 3/17 & 17/3 & 20/0 & 0/20 & 16/4 & 18/2 & 1/19 & 18/2 & 20/0 \\
Qwen3.5-9B & 9/11 & 8/12 & 13/7 & 13/7 & 13/7 & 14/6 & 11/9 & 13/7 & 16/4 \\
aya-expanse-32b & 0/20 & 17/3 & 5/15 & 0/20 & 15/5 & 11/9 & 0/20 & 20/0 & 20/0 \\
gemma-4-26B-A4B-it & 0/20 & 20/0 & 20/0 & 0/20 & 20/0 & 20/0 & 0/20 & 20/0 & 20/0 \\
gemma-4-31B-it & 0/20 & 20/0 & 20/0 & 0/20 & 20/0 & 20/0 & 0/20 & 20/0 & 20/0 \\
gemma-4-E4B-it & 13/7 & 20/0 & 20/0 & 9/11 & 20/0 & 18/2 & 10/10 & 20/0 & 20/0 \\
gpt-oss-20b & 0/20 & 11/9 & 16/4 & 0/20 & 10/10 & 13/7 & 0/20 & 14/6 & 17/3 \\
phi-4 & 0/20 & 10/10 & 17/3 & 0/20 & 4/16 & 12/8 & 0/20 & 15/5 & 20/0 \\
\midrule
\multicolumn{10}{l}{\textbf{Language: German}} \\
\midrule
Llama-3.1-8B-Instruct & 2/18 & 8/12 & 13/7 & 2/18 & 5/15 & 11/9 & 0/20 & 7/13 & 13/7 \\
Llama-3.3-70B-Instruct & 0/20 & 20/0 & 20/0 & 0/20 & 19/1 & 20/0 & 0/20 & 20/0 & 20/0 \\
Qwen3.5-27B & 14/6 & 20/0 & 20/0 & 5/15 & 18/2 & 20/0 & 14/6 & 20/0 & 20/0 \\
Qwen3.5-35B-A3B & 0/20 & 17/3 & 14/6 & 0/20 & 14/6 & 12/8 & 0/20 & 14/6 & 19/1 \\
Qwen3.5-9B & 6/14 & 12/8 & 13/7 & 3/17 & 12/8 & 9/11 & 8/12 & 13/7 & 14/6 \\
aya-expanse-32b & 0/20 & 0/20 & 2/18 & 0/20 & 0/20 & 0/20 & 0/20 & 0/20 & 2/18 \\
gemma-4-26B-A4B-it & 0/20 & 20/0 & 20/0 & 0/20 & 20/0 & 20/0 & 4/16 & 20/0 & 20/0 \\
gemma-4-31B-it & 0/20 & 20/0 & 20/0 & 0/20 & 20/0 & 20/0 & 0/20 & 20/0 & 20/0 \\
gemma-4-E4B-it & 1/19 & 20/0 & 12/8 & 3/17 & 13/7 & 9/11 & 1/19 & 20/0 & 20/0 \\
gpt-oss-20b & 0/20 & 12/8 & 9/11 & 0/20 & 8/12 & 11/9 & 1/19 & 16/4 & 13/7 \\
phi-4 & 3/17 & 20/0 & 20/0 & 1/19 & 19/1 & 20/0 & 8/12 & 20/0 & 20/0 \\
\midrule
\multicolumn{10}{l}{\textbf{Language: Spanish}} \\
\midrule
Llama-3.1-8B-Instruct & 2/18 & 6/14 & 6/14 & 1/19 & 6/14 & 4/16 & 4/16 & 8/12 & 12/8 \\
Llama-3.3-70B-Instruct & 0/20 & 20/0 & 20/0 & 0/20 & 20/0 & 20/0 & 0/20 & 20/0 & 20/0 \\
Qwen3.5-27B & 11/9 & 20/0 & 20/0 & 13/7 & 20/0 & 20/0 & 9/11 & 20/0 & 20/0 \\
Qwen3.5-35B-A3B & 6/14 & 18/2 & 20/0 & 5/15 & 15/5 & 20/0 & 9/11 & 16/4 & 20/0 \\
Qwen3.5-9B & 12/8 & 17/3 & 18/2 & 16/4 & 15/5 & 17/3 & 12/8 & 19/1 & 17/3 \\
aya-expanse-32b & 20/0 & 20/0 & 20/0 & 20/0 & 20/0 & 20/0 & 20/0 & 20/0 & 20/0 \\
gemma-4-26B-A4B-it & 0/20 & 20/0 & 20/0 & 0/20 & 20/0 & 20/0 & 0/20 & 20/0 & 20/0 \\
gemma-4-31B-it & 0/20 & 20/0 & 20/0 & 0/20 & 20/0 & 20/0 & 0/20 & 20/0 & 20/0 \\
gemma-4-E4B-it & 0/20 & 7/13 & 7/13 & 0/20 & 5/15 & 7/13 & 0/20 & 6/14 & 15/5 \\
gpt-oss-20b & 1/19 & 12/8 & 11/9 & 0/20 & 9/11 & 11/9 & 0/20 & 12/8 & 14/6 \\
phi-4 & 9/11 & 20/0 & 20/0 & 8/12 & 20/0 & 20/0 & 14/6 & 20/0 & 20/0 \\
\bottomrule
\end{tabular}
\caption{Consistency results (Consistent/Adaptation) across all languages, countries, and professions.}
\label{tab:consistency_all}
\end{table*}
\end{document}